\documentclass[final]{clear2023} 

\title[Evaluating Temporal Observation-Based Causal Discovery Techniques]{Evaluating Temporal Observation-Based Causal Discovery Techniques Applied to Road Driver Behaviour}
\usepackage{times}

\clearauthor{\Name{Rhys Howard} \Email{rhyshoward@live.com}\and
 \Name{Lars Kunze} \Email{lars@robots.ox.ac.uk}\\
 \addr Oxford Robotics Institute, Dept. of Engineering Science, University of Oxford, 17 Parks Road, Oxford,\\ OX1 3PJ, United Kingdom}

\usepackage{csvsimple} 
\usepackage{comment}

\begin{document}

\maketitle

\begin{abstract}%
Autonomous robots are required to reason about the behaviour of dynamic agents in their environment. The creation of models to describe these relationships is typically accomplished through the application of causal discovery techniques. However, as it stands observational causal discovery techniques struggle to adequately cope with conditions such as causal sparsity and non-stationarity typically seen during online usage in autonomous agent domains. Meanwhile, interventional techniques are not always feasible due to domain restrictions. In order to better explore the issues facing observational techniques and promote further discussion of these topics we carry out a benchmark across 10 contemporary observational temporal causal discovery methods in the domain of autonomous driving. By evaluating these methods upon causal scenes drawn from real world datasets in addition to those generated synthetically we highlight where improvements need to be made in order to facilitate the application of causal discovery techniques to the aforementioned use-cases. Finally, we discuss potential directions for future work that could help better tackle the difficulties currently experienced by state of the art techniques.
\end{abstract}

\begin{keywords}%
  Causal Discovery, Time Series Data Analysis, Autonomous Driving
\end{keywords}

\section{Introduction}

With an increase in robots operating among humans, there is a growing need for these autonomous agents to understand how the actions of one agent - human or robotic - may affect the behaviour of other agents. In this paper we present a benchmark evaluation of contemporary temporal causal discovery methods in attempting to find causal relationships between agents. 

The incentive for building a causal model of agent behaviours is to allow consideration of humans or other robots when planning, and then being able to retrospectively explain plans and outcomes \citep{hellstrom2021relevance}. Accounting for causality has also been identified as a critical component in developing trustworthy AI systems \cite{ganguly2023review}.
This type of work has become increasingly relevant due to the rise in robots operating in close proximity to humans, particularly in high risk domains such as driving. Human drivers must constantly be aware of other vehicles and act in a way as to not only achieve their objectives but simultaneously be aware of the consequences of their actions, both direct and indirect. By furthering discussion in this area we aim to encourage work towards developing a similar reasoning model for AI and robotics. 
\begin{figure}[t]
    \centering
    \subfigure[Initial frame.]{
        \includegraphics[width=0.31\textwidth]{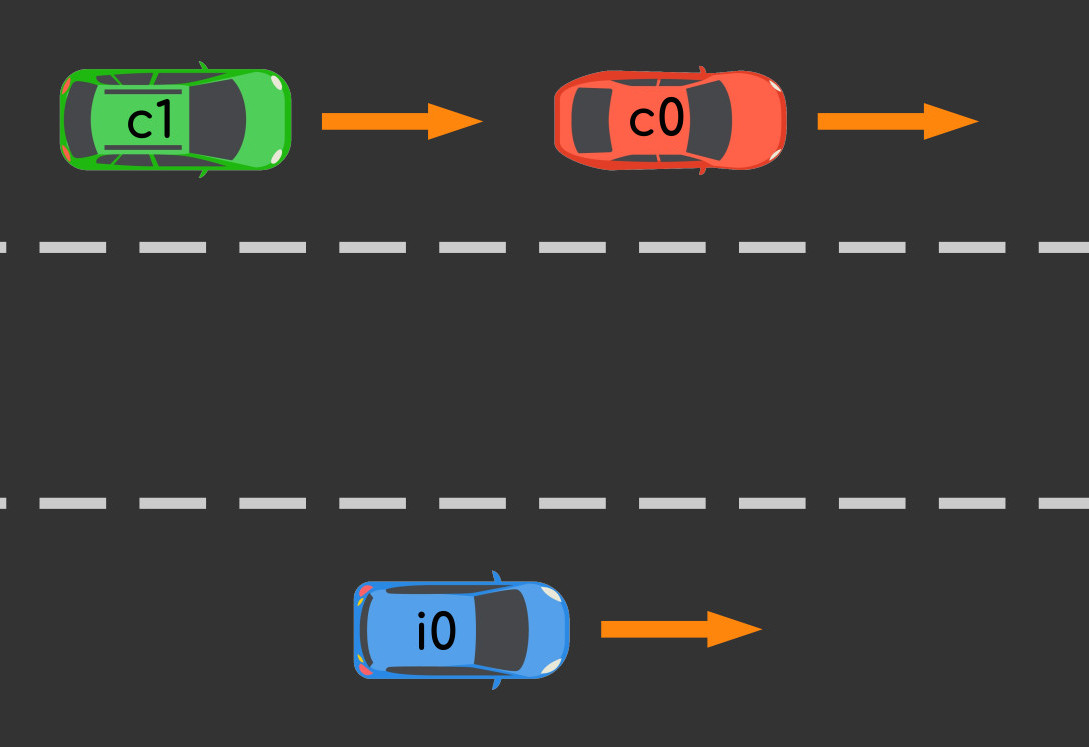}
    } \label{subfig:two_car_convoy_a}
    \subfigure[Post-\textbf{c0} braking.]{
        \includegraphics[width=0.31\textwidth]{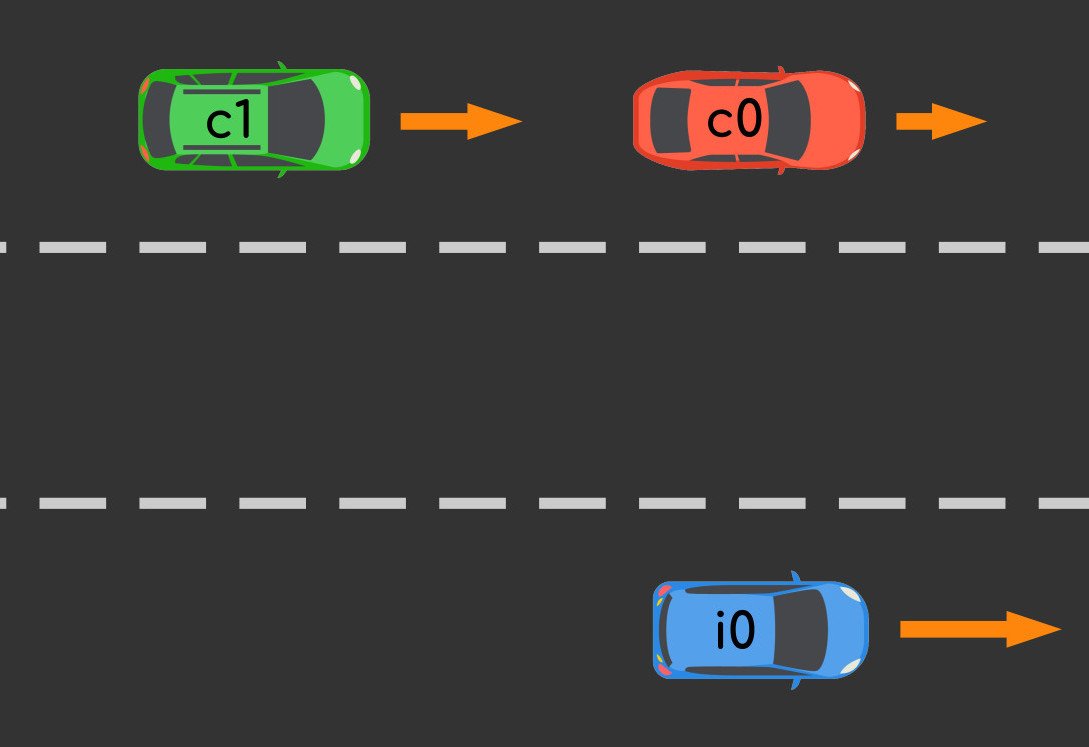}
    } \label{subfig:two_car_convoy_b}
    \subfigure[Causal graph.]{
        \includegraphics[width=0.31\textwidth]{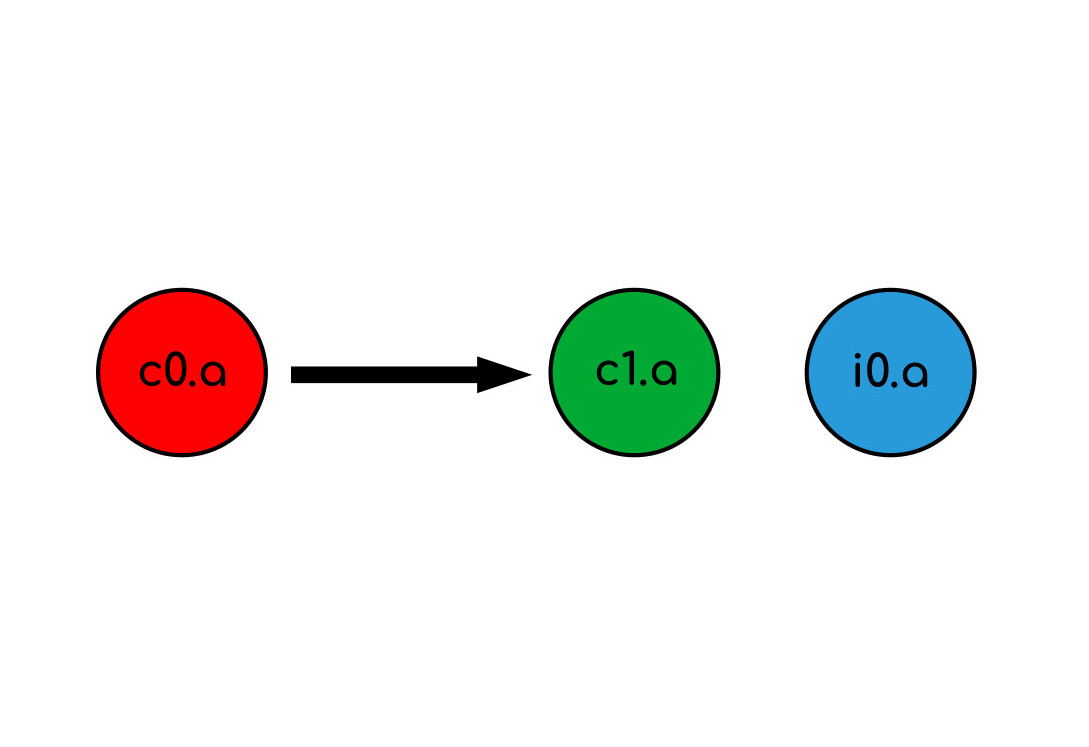}
    } \label{subfig:two_car_convoy_c}
    \caption{Illustration of the basic causal interaction we consider: The two-agent convoy scenario. Here \textbf{c0} and \textbf{c1} represent the head and tail of a two vehicle convoy respectively, while \textbf{i0} acts as an entirely independent third agent. \citep{freepik2022cars}}
    \label{fig:two_car_convoy}
\end{figure}


In order to explore the best paths forward for developing a system of causal discovery for agent behaviour the state of the art must first be evaluated. The methods covered in this work consist of those presented by \citet{assaad2022survey}, the NAVAR method presented by \citet{bussmann2021neural} and the CD-NOD method presented by \citet{zhang2017causal}. In contrast to this work, the work of Assaad et al. only evaluates the methods on artificial data generated from common causal graph structures and fMRI simulation \citep{smith2011network}.

Working in an online fashion to establish causal links between agent behaviour presents a few challenges for the aforementioned methods. Most applications of causal discovery are upon problems with large amounts of available data where processing can be carried out in an offline setting (e.g. medicine, economics, sociology). Due to the quantity and nature of the data, there are likely to be a greater number of causal interactions resulting from causal relationships. However if data is processed in an online fashion and causal interactions indicative of causal relationships are infrequent, this leads to a condition we term causal sparsity. Under this condition, causal discovery methods must contend with being able to carry out discovery with substantially fewer observations of causal relationships being exhibited than would usually be the case. Most causal discovery tasks also assume causal stationarity in the ground truth causal relationships, either due to the synthetically generated data or due to the very nature of the problem being considered. Causal stationarity refers to causal relationships that remain constant throughout the observed time window. Because we consider human drivers, which vary in their reaction times and in the manner of their reactions, non-stationarity in their causal relationships is inevitable. 

With the motivation, content and challenges in relation to this work laid out, we identify the following as the key contributions of this work:
\begin{itemize}
    \item A benchmark evaluation of contemporary temporal causal discovery techniques applied to real world scenes which follow a predetermined Autonomous Driving (AD) scenario (See Sec. \ref{sec:problem_definition}).
    \item A comparison of real world scene performance with performance on scenes synthetically generated with ideal conditions.
    \item A discussion of the difficulties associated with causal discovery on agent behavioural interactions on a scene-by-scene basis, and the limitations presented by some methods as a result.
\end{itemize}
We note that while this paper does focus upon an AD scenario, that the challenges encountered while tackling this problem are likely shared by other areas of autonomous robotics, particularly those where an interest in reasoning about agent behavioural interaction is present.



\section{Problem Definition} \label{sec:problem_definition}

To the author's knowledge the methods under consideration have not had their performance tested in the domain of AD before. Therefore we intend to carry out our investigation using a simple scenario that aims to establish the viability of the techniques in question. 
The types of causal relationships we are considering here are behavioural and not physical interactions between agents as the intended application of the discovered causal models is for decision explanation and introspection. In order to consider two agents as causally linked the behaviour of the causing agent must either be sufficient or necessary to produce some of the behaviour exhibited by the effected agent.

The scenario in question describes two primary agents \textbf{c0} and \textbf{c1} which form the head and tail of a two vehicle convoy. Throughout the course of the scenario, \textbf{c0} will at one point noticeably accelerate or decelerate and in response, \textbf{c1} will accelerate or decelerate accordingly, forming the basis of a causal relationship between the two agents. With the aim of demonstrating the precision of the approaches, we introduce a third independent agent \textbf{i0}. As the name suggests, the behaviour of \textbf{i0} is entirely independent of \textbf{c0} and \textbf{c1}, and as a result there should be no causal relationships detected between the convoy agents and \textbf{i0}. An illustration of this scenario is shown in Fig. \ref{fig:two_car_convoy}.

We can prove that either an acceleration or deceleration case shows a valid causal link between agents by considering the alternate scenarios in which the lead vehicle did not accelerate/decelerate. In the deceleration case, as it is clear that that \textbf{c0} braking is sufficient to cause \textbf{c1} to brake, as \textbf{c1} would collide with \textbf{c0} otherwise. We can also consider the acceleration of \textbf{c0} necessary in causing \textbf{c1} to accelerate, as \textbf{c1} could not accelerate to a speed greater than \textbf{c0} without risking collision. 

In terms of the numeric variables used in the causal discovery process, all three agents will rely purely either on their linear acceleration magnitudes denoted as \textbf{c0.a}, \textbf{c1.a}, and \textbf{i0.a} respectively or on their linear velocity magnitudes again denoted \textbf{c0.v}, \textbf{c1.v}, and \textbf{i0.v} respectively. 

Finally, while we will not refrain from working with egocentric data, the scenario described above is not focused upon an ego agent/vehicle. Here the use of the term ego refers to data gathered or problems considered from a single agent/vehicle's perspective. In contrast to previous works \citep{kim2017interpretable,li2020who} which have looked at causal discovery from a single point of view, we instead aim to investigate the described scenario in a perspective-agnostic manner.


\section{Related Work} \label{sec:related_work}

\subsection{Causal Reasoning}

Causal reasoning as a well defined area of study spawned out of work by \citet{pearl2009causality}, expanding upon Bayesian networks \citep{pearl1985bayesian}. Causal models provide information regarding the generative processes involved in the creation of data, rather than just the associations between variables/events. Causal discovery is the process of inferring causal models, and in the context of AD we hope to discover causal links between the behaviour of road agents.

Causal discovery can be carried out at one of three layers \citep{pearl2019book}: Observational, Interventional, or Counterfactual. Interventional methods which inherently require experimentation are unsuitable for the AD domain we consider \citep{eberhardt2007interventions,kocaoglu2017experimental,addanki2020efficient}. Meanwhile, counterfactual methods while interesting theoretically, have had little attention applied to them thus far outside of a meteorological simulation based approach relating to climate change \citep{hannart2016causal}. As such, in this paper we focus on the application of contemporary observational approaches to the AD domain.

\subsection{Causal Discovery on Time Series Data}

\citet{glymour2019review} provides a description of contemporary observational causal discovery methods as well as discussing several difficulties typically encountered by these methods, including the extension of said methods to time series data. Expanding upon this \citet{assaad2022survey} conducts a survey of observational approaches applied to time series data, and provides a quantitative evaluation of the approaches on synthetic and real data. The evaluated methods in question are the Granger causality based PWGC \citep{granger1969investigating}, MVGC \citep{geweke1982measurement}, and TCDF \citep{nauta2019causal} approaches, the constraint based PCMCI \citep{runge2019detecting}, oCSE \citep{sun2015causal}, and tsFCI \citep{entner2010causal} approaches, the noise based VarLiNGAM \citep{hyvarinen2010estimation} and TiMINo \citep{peters2013causal} approaches, and the score based DYNOTEARS \citep{pamfil2020dynotears} approach. Of these approaches oCSE is not relevant to this paper, as it assumes causal relationships only occur over a time lag of 1, which is far too little for the task we consider. In general, the methods show limited performance when self-causal relationships are excluded, which contributes towards our selection of a simple scenario to evaluate the approaches upon.
In addition to the methods reviewed by Assaad et al., we also consider the Granger causality based NAVAR method presented by \citet{bussmann2021neural} and the non-stationarity-based method CD-NOD presented by \citet{zhang2017causal}. NAVAR takes a similar approach to TCDF \citep{nauta2019causal} in that it applies neural learning techniques to the causal discovery problem, meanwhile CD-NOD bears some resemblance to traditional constraint-based methods, but additionally exploits non-stationarity to discover causal direction.

\subsection{Causal Discovery in the Autonomous Driving Domain}

While works combining causal discovery in the domain of AD are rare, there have been a few recent works which focus on this area. One such work is that carried out by \citet{mcduff2021causalcity}. They present a detailed simulation environment to facilitate future causal research, and contrast this environment with a drastically simplified environment to demonstrate the increased challenge and realism that comes with the detail of the approach they describe. In their work they test their simulated scenarios against three causal discovery techniques.

Of the three methods evaluated, the NRI method \citep{kipf2018neural} operates on 2D particle systems and motion capture data, the NS-DR method \citep{yi2020clevrer} presents itself by testing on 3D particle systems, and the V-CDN method \citep{li2020causal} consists of an entire pipeline dedicated to detecting key points on clothing and determining the physical relationships between said points with a causal model. All three of these methods can be considered visual causal discovery in that they are restricted to working upon visual input. Furthermore, due to all three of them utilising neural networks it increases the difficulty of performance verification and increases the risk of performance degradation from domain shift.

In terms of other works tackling causal discovery in the domain of AD, \citet{kim2017interpretable} and \citet{li2020who} both attempt to discover which regions of a camera's view are responsible for invoking certain ego agent behaviours. The main limitation of such an approach is that it is limited to ego vehicles utilising cameras, cannot reason about the actions of other agents, and once again relies upon convolutional neural network techniques. Meanwhile \citet{dehaan2019causal} and \citet{samsami2021causal} both utilise causal discovery as part of an imitation learning process for AD control, as opposed to our goal of evaluating behavioural interactions between agents. 


\section{Background}

A causal directed graph is defined as $G = (V, E)$ where $V$ corresponds to a set of variables and $E \subseteq [ (v^x, v^y)\ |\ v^x \neq v^y, (v^x, v^y) \in V \times V]$ corresponds to a set of edges that describe causal relationships between the aforementioned variables. An edge $(v^x, v^y) \in E$ describes a causal relationship whereby variable $v^x$ has a causal effect upon variable $v^y$. Causal Directed Acyclical Graphs (DAGs) comprise a subset of causal directed graphs that are without any cycles, a feature which is exploited by some causal discovery approaches when discovering DAGs. If the variables being considered consist of time series, a subscript is added to denote the variable at the specified time. For example, the variable $v^x \in V$ is denoted at time $t$ as $v^x_t$.

The goal of causal discovery is to derive an approximation of the causal directed graph $\hat{G} = (\hat{V}, \hat{E})$. Here $\hat{E}$ is simply the hypothesised causal links, meanwhile $\hat{V}$ consists of the approximated set of variables, as there can exist the possibility of hidden confounders, which some causal discovery approaches make efforts to detect. It should be noted that while an assumption of causal sufficiency ensures that $\hat{V} = V$, we cannot rule out the presence of hidden confounders. The potential implications of this will be discussed in more detail in Sec. \ref{sec:experiments}, but for the purposes of the discussing methodology only the three variables associated with each respective agent will ever be considered as part of our evaluation.

In terms of the problem defined in Sec. \ref{sec:problem_definition}, we are looking to discover a causal temporal summary graph, in which there is an edge directed from the time series relating to agent \textbf{c0} to the time series relating to the agent \textbf{c1} and none besides that (See Fig. \ref{fig:two_car_convoy}). A summary graph just means that provided there exists a causal relationship between two time series within a given time window given by $\tau$ an edge will be present in the summary graph to represent said relationship.


\section{Observation-Based Temporal Causal Discovery}

Here the we will briefly introduce the methods under consideration based upon a selection of those presented by \citet{assaad2022survey} in their survey paper, in addition to the NAVAR method \citep{bussmann2021neural}. We refer the reader to their papers for more substantial explanations of the methods paraphrased here.

\subsection{Granger Causality Based Approaches}

Granger causality is built upon the concept that causes should provide unique information that enables the prediction of their effects \citep{granger1969investigating}. The first two Granger causality based approaches assume a linear relationship between variables, which should indeed be true of the variables we consider, as the tail convoy vehicle should roughly mirror the speed and acceleration of the head convoy vehicle, albeit with a time lag. The Pair-Wise Granger Causality (PWGC) \citep{granger1969investigating} approach determines the likelihood of a causal link between variables by considering two autoregressive models:
\begin{equation} \tag{PW-res.} \label{eq:pwres}
\small
    v^y_{t} = \sum^{\tau}_{t^\prime = 1} (a^y_{t^\prime} v^y_{t - t^\prime}) + {\varepsilon}^y_t
\end{equation}
\begin{equation} \tag{PW-full} \label{eq:pwfull}
\small
    v^y_{t} = \sum^{\tau}_{t^\prime = 1} (a^y_{t^\prime} v^y_{t - t^\prime} + a^x_{t^\prime} v^x_{t - t^\prime}) + {\varepsilon}^y_t
\end{equation}
where $a^x_{t^\prime}, a^y_{t^\prime} \in \mathbb{R}$ represent coefficients specific to a time lag $t^\prime$ for variables $v^x$ and $v^y$ respectively and $\tau$ is the maximum tag lag considered. Meanwhile ${\varepsilon}^y$ and ${\varepsilon}^y$ are white-noise time series which represent the influence of exogenous non-confounding factors.
If $v^x$ does indeed have a causal effect upon $v^y$ we would expect $v^x$ to possess unique information allowing us to better predict $v^y$ with (\ref{eq:pwfull}) than with (\ref{eq:pwres}). An F-test can be applied to an accuracy metric (e.g. Residual Sum of Squares) for each of the models to determine whether the difference is significant enough to consider a causal link to have been discovered.

The PWGC approach is limited in that it only ever considers pairs of variables and can therefore struggle with mediators relationships present in the underlying causal graph. Multi-Variate Granger Causality (MVGC) \citep{geweke1982measurement} builds upon PWGC by proposing the following autoregressive models at the cost of higher computational overhead:
\begin{equation} \tag{MV-res.} \label{eq:mvres}
\small
    v^y_{t} = \sum_{v^i \in V^{\neg x}} (\sum^{\tau}_{t^\prime = 1} (a^i_{t^\prime} v^i_{t - t^\prime})) + {\varepsilon}^y_t
\end{equation}
\begin{equation} \tag{MV-full} \label{eq:mvfull}
\small
    v^y_{t} = \sum_{v^i \in V} (\sum^{\tau}_{t^\prime = 1} (a^i_{t^\prime} v^i_{t - t^\prime})) + {\varepsilon}^y_t
\end{equation}
where $V^{\neg x} = V \setminus \{v^x\}$. Unlike PWGC, which only considers self causation in the restricted case and the additional information provided by a single variable in the full case, MVGC considers all bar one variable in (\ref{eq:mvres}) and all variables in (\ref{eq:mvfull}). Because this approach considers all information available while only excluding the information of one variable that is being examined as a cause, it is not only able to tackle mediator causal relationships, but is also robust against against confounding variables as their influence is captured in both models with equal measure. 

The Temporal Causal Discovery Framework (TCDF) \citep{nauta2019causal} and Neural Additive Vector Auto-Regression (NAVAR) \citep{bussmann2021neural} approach both offer neural network based approaches to model non-linear causal relationships. TCDF utilises a form of attention-based convolutional neural network to predict for a given variable provided the historic values of all variables, the attention values attributed during training are then used to determine potential causal parents for the given variable. Meanwhile NAVAR trains a network - a Mutli-Layer Perceptron (MLP) in our experiments - for each variable to predict for all other variables, in this case relying upon the variance a variable provides in its prediction contributions to determine the presence of causal relationships.

\subsection{Noise-Based Approaches}

Noise-based approaches share a theoretical similarity with Granger causality based approaches in that these approaches also rely upon examining the flow of information between variables. However, in contrast to Granger causality, noise-based methods do not operate upon identifying information possessed by variables useful for predicting other variables. Instead they attempt to identify the direction of causal links by identifying information variables hold about the noise of other variables.

To explain the premise of the first noise-based method, VarLiNGAM \citep{hyvarinen2010estimation}, consider the following bivariate example of LiNGAM \citep{shimizu2006linear} its non-temporal predecessor. Let $v^x$ have a causal effect upon $v^y$ and that their distribution be generated as follows:
\begin{equation}
\small
    v^x = {\varepsilon}^x
\end{equation}
\begin{equation}
\small
    v^y = a^{x,y} v^x + {\varepsilon}^y
\end{equation}
where ${\varepsilon}^x$ and ${\varepsilon}^y$ represent noise from exogenous non-confounding factors that affect the variables. Under this distribution, $v^y$ captures information on the noise provided by ${\varepsilon}^x$ because its value is derived from $v^x$. However, $v^x$ does not capture any information on ${\varepsilon}^y$, establishing an asymmetry that LiNGAM aims to exploit. It is important to note that ${\varepsilon}^x$ and ${\varepsilon}^y$ must not be jointly Gaussian, as under these conditions it is impossible to determine the causal direction. 

In order to solve the directions of causal links the model reframes the underlying model responsible for the time series as $V = AV + \varepsilon$ where $V$ is a vector of variables, $\varepsilon$ is a vector of noise/error values, and $A$ is a strictly lower triangular matrix which describes the direction of causal links between the variables of $V$. If the model is further refined to $V = B\varepsilon$ where $B=(I-A)^{-1}$ we can aim to try and solve for $A$. The initial paper on LiNGAM \citep{shimizu2006linear} applied Independent Component Analysis (ICA) \citep{comon1994independent} in order to achieve this. However the method explored by this paper utilises an extension named DirectLiNGAM \citep{shimizu2011directlingam}.
This method constructs an auto-regressive model and recursively checks independence between each variable acting as a predictor and the residuals given by applying that predictor to other variables. The most independent predictor is placed highest in the causal hierarchy. In each subsequent step the remaining variables are substituted for their residuals obtained during the previous step to remove the influence of variables with established positions. Since the above process only establishes the direction of causation, the strength of the causal effect can be determined by conventional covariance-based regression, before pruning is carried out by applying the Adaptive Lasso method \citep{zou2006adaptive}. From here the extension to VarLiNGAM \citep{hyvarinen2010estimation} is achieved by considering the variables in question over a time window defined by the maximum time lag $\tau$:
\begin{equation}
\small
    V_{t} = \sum^{\tau}_{t^\prime = 1} (A_{t^\prime} V_{t - t^\prime}) + {\varepsilon}_t
\end{equation}
Provided one can approximate $(A_{t^\prime})_{1\geq t^\prime \geq \tau}$ the steps of DirectLiNGAM can be followed in similar fashion albeit working with time series rather than regular non-temporal variables. The use of the Adaptive Lasso remains important as prior to this step the number of causal links will be of the order $\mathcal{O}({|V|}^2\tau)$, however the risk for our problem at least more concerns the potential for the inclusion of false positives in the summary graph.

The other noise-based approach, TiMINo \citep{peters2013causal} consists not so much of a single model as much as it describes a class of models. The models described by TiMINo are assumed to adhere to the form:
\begin{equation}
\small
    v^x_t = f^x (pa(v^x_t, t), pa(v^x_t, t - 1), ..., pa(v^x_t, t - \tau), \varepsilon^x_t)
\end{equation}
where $pa(v^x_t, t^\prime)$ represents the causal parents of $v^x_t$ from the time $t^\prime$ up to maximum time lag given by $\tau$. The function $f^x$ and additive noise $\varepsilon^x_t$ are partially dependent upon one another, as if $f^x$ is non-linear, $\varepsilon^x_t$ should be Gaussian, and if $f^x$ is linear, $\varepsilon^x_t$ should be non-Gaussian. Alternatively these requirements can be relaxed if the data follows a time structure - as opposed to independent and identically distributed time indexed variables - that the joint data distribution is faithful to and there is a lack of cycles. Due to the nature of the data being worked with, this latter case captures our task more closely. In terms of how TiMINo operates, it utilises a supplied regression method and independence test. The implementation utilised by this paper \citep{assaad2022survey} utilises a linear regression model and a cross covariance based independence test with Bonferroni correction. TiMINo proceeds by learning a predictor for each time series and then determining how independent each time series involved in the predictor is from the residuals produced from applying the predictor. The time series with the predictor that produces the greatest level of independence is deemed to be at the bottom of the causal hierarchy as little or no information regarding its own noise is present in other time series. This process is repeated until a full causal hierarchy is established. At this point the causal parents of each time series are refined by removing those time series unnecessary to produce independent residuals.

\subsection{Constraint-Based Approaches}

Constraint-based approaches make the assumption that the conditional independences seen in the probability distributions exhibited by observed data are reflective of the underlying causal graph structure, a condition referred to as causal faithfulness. Constraint-based approaches typically work upon the time series considered in the form of a window causal graph, where each relative time lag - time series combination represents a node within the graph. Both of the constraint-based methods we consider share a similar initial process of first identifying which which nodes are unconditionally independent before progressively checking independence upon adjacent node pairs conditional upon their neighbours. Once adjacencies remain stable one can look for unshielded triples and convert these into collider structures as these are the only structures that can be directly identified under the assumptions of the methods considered. 

One of the oldest constraint-based approaches is the Peter-Clark (PC) algorithm \citep{spirtes2001causation}. Following the steps outlined above the PC algorithm proceeds to iteratively apply 3 rules for refining the causal direction of adjacent nodes. Depending upon the data available it will not always be possible to orient every edge. The resulting graph describes a Markov Equivalence Class (MEC) which contains all the possible full directed causal graphs based upon the remaining undirected edges, should any be present. The PC with Momentary Conditional Independence (PCMCI) \citep{runge2019detecting} extends the PC algorithm to better work with time series. While the initial steps are similar to the PC algorithm, the Momentary Conditional Independence test is designed to avoid the influence of auto-correlations by evaluating the level of dependence between nodes while conditioning upon the parents of both nodes. It is important to note that PCMCI can be used with any conditional independence test, though for this paper only the partial correlation \citep{baba2004partial} approach is considered. 

The other constraint-based method, tsFCI \citep{entner2010causal} is based upon the earlier FCI algorithm \citep{zhang2008completeness}. The FCI follows the same initial process as the PC algorithm, however when it comes to the iterative application of rules there are 10 rules as opposed to 3. The purpose of these additional rules to allow FCI to accommodate the presence of exogenous hidden confounder ancestors and hidden descendants that have been inadvertently conditioned upon (e.g. selection bias). To achieve this FCI works upon the concept of a Maximal Ancestral Graph (MAG) rather than a MEC, that allows for bi-directional edges representing hidden confounder ancestors, and non-directional edges representing conditioned upon hidden descendants. 

\subsection{Non-Stationarity-Based Approaches}
While most approaches struggle with non-stationarity, approaches such as CD-NOD proposed by \citet{zhang2017causal} actively aim to exploit non-stationarity. For the most part CD-NOD resembles a constraint-based approach except with the addition of a surrogate variable designed to capture either domain-based or time-based distribution shifts. The discovery of the causal graph skeleton is carried out in a similar manner to a method such as PC, relying upon conditional independence tests and applying orientation rules. However, in determining causal direction, CD-NOD bears some similarities to a noise-based approach, except rather than examining the propagation of noise CD-NOD examines the influence of variable specific parameters calculated as a function of the surrogate variable. Since the actual parameters are unknown CD-NOD tackles the problem through the application of Kernel Density Estimation (KDE). One limitation of this approach is that if the surrogate variable has control over a confounding variable, CD-NOD is reliant upon the influence of the confounding variable being weak compared to the surrogate controlled variable parameters.

In the aforementioned work, CD-NOD was purely intended for application on instantaneous causal relationships, however it was discussed that it was natural to extend the method to include time lagged relationships, this was subsequently accomplished by \citet{huang2020causal}. However a shortcoming of this approach as a whole in the context of this work, is while CD-NOD can capture shifts in causal model parameterisation, it cannot capture changes to the causal skeleton that occur with time, nor can it capture shifts in causal direction with time. This is also true of the time lagged extension, meaning that a causal relationship that varies in time lag over time but not with causal effect cannot be adequately captured. Nevertheless of the approaches evaluated in this paper, CD-NOD offers the best framework to provide robustness against non-stationarity.

\subsection{Score-Based Approaches}

Score-based approaches view the causal discovery process as the task of finding a causal graph which maximises a scoring metric of how well the data fits the supplied graph. Since an exhaustive search would be computationally intractable this is typically tackled as an optimisation problem. In terms of the score metrics used for this type of approach, the Bayesian Information Criterion (BIC) \citep{schwarz1978estimating}, Bayesian Dirichlet equivalence (BDe) \citep{heckerman1995learning} score, and Cross-Validation (CV) \citep{pena2005learning} technique have all been proposed as options.

However, the approach we consider for evaluation, DYNOTEARS \citep{pamfil2020dynotears} defines a new metric for this purpose. The following consists of the metric they define without any of the components relating to instantaneous causal relationships and considering a single independent realization of each underlying causal model, as this better reflects the problem we consider:
\begin{equation} \label{eq:dynotears_score}
\small
    f(A) = \ell(A) + \lambda_A \max_{v^j \in V} \sum^{\tau}_{t^\prime=1} \sum_{v^i \in V} (a^{i,j}_{t^\prime})
\end{equation}
\begin{equation} \label{eq:dynotears_loss}
\small
    \ell(A) = \frac{1}{2n} \sum^{t_{max}}_{t=\tau} \sum_{v^i \in V} (v^i_t - \sum^{\tau}_{t^\prime=1} \sum_{v^j \in V} (a^{i,j}_{t^\prime} v^j_{t - t^\prime}))
\end{equation}
where $n=t_{max} + 1 - \tau$, $t_{max}$ is the largest time index present in the time series, $\lambda_A$ is a regularisation constant, and $a^{i,j}_{t^\prime}$ corresponds to the $i$-th row and $j$-th column of $A$ at the time lag of $t^\prime$. Here (\ref{eq:dynotears_loss}) describes a loss term derived from the difference between the time series values predicted by $A$ and the observed time series values, meanwhile (\ref{eq:dynotears_score}) reflects an overall score derived from combining (\ref{eq:dynotears_loss}) with a regularisation term. After optimising $A$ we can then construct a causal graph by thresholding each element $a^{i,j}_{t^\prime}$ within $A$, and adding a graph edge should the value prove great enough. As a result $A$ reflects both a weight matrix updated while learning a linear predictive model with DYNOTEARS, and an adjacency matrix which is later converted to a causal graph by applying a threshold to its elements.


\section{Experiments} \label{sec:experiments}

\subsection{Datasets \& Code} \label{subsec:datasets_and_code}

The experiments were carried out on three datasets independently, the Lyft Level 5 Prediction dataset, the High-D dataset and a synthetic dataset. All the scenes in this dataset exhibit the causal relations/interactions described in Sec. \ref{sec:problem_definition}. We make the code used to carry out these experiments as well as the data associated with them that can be publicly shared available in a Git repository\footnote{https://github.com/cognitive-robots/temporal-cd-evaluation-paper-resources}.

\subsubsection{Lyft Level 5 Prediction}

This dataset is comprised of over a thousand hours of driving data collected over 20 autonomous vehicles operating in Palo Alto, California \citep{houston2020one}. The data itself is comprised of egocentric data (e.g. position/orientation of capturing autonomous vehicle), labeling of vehicles and pedestrians complete with position, orientation and bounding boxes, and the status of any perceived traffic lights. This time series data is structured into a series of scenes, each ${\sim}\,30\ s$ in length and captured at $10\ Hz$. In addition to these scenes, there is a static semantic map describing the road network the autonomous vehicles were operating on.

In order to transform the data available into two-agent convoy scenes upon which causal discovery could be carried out, a manual inspection was carried out upon each scene in turn. This was used to determine which scenes exhibited the scenario laid out in Sec. \ref{sec:problem_definition}, and within these scenes which agents - besides the ego vehicle - would form part of the scene. In total 50 scenes were extracted from the dataset in this manner, all of which were at least $10\ s$ in length following any trimming of the scene that needed to be carried out. 
Finally, the velocities and accelerations of agents were estimated based upon the change in agent positions between frames, and in order to account for positional jitter, the velocity and acceleration were smoothed via a 15 frame moving average.

\subsubsection{High-D}

In contrast to the variety of roads within Palo Alto covered by the Lyft dataset, the High-D dataset focuses upon stretches of highway at six locations across Germany \citep{krajewski2018highd}. It offers greater coverage of the roads it considers by capturing overhead footage using a drone. This is automatically annotated, resulting in a positional error of labelled vehicles typically under $10\ cm$. The recordings that make up the dataset were captured at $25\ Hz$ over the course of several minutes. As opposed to a detailed semantic map, the dataset simply offers the $y$ position of the lane markings in pixel space, as all data is aligned to have the lanes run parallel to the $x$ axis. Unlike the Lyft dataset, the High-D dataset also provides stable velocity and acceleration values out of the box, and as such these did not need to be calculated. The data was however resampled to $10\ Hz$ as to maintain a similar maximum time lag in seconds.

Due to additional details in terms of lane occupancy and inter-vehicular adjacency provided by the dataset, it was possible to automate the process of creating scenes by simply searching the dataset for instances where this convoy-like causal relationship occurs and then selecting a third vehicle from another lane to act as \textbf{i0}. Once again, all produced scenes were of at least $10\ s$ in length, 
though most were longer due to the wide area captured by the drone camera.
This led to a total of 3395 scenes being extracted.

\subsubsection{Synthetic}

In addition to running experiments upon the two real world datasets we also generated and ran experiments upon a synthetic dataset. Doing so facilitates differentiating between performance loss due to the nature of the scenario from performance loss due to real world complications (e.g. sensor noise, variations in human behaviour, etc.). In order to generate this dataset, a series of velocity goal objectives are assigned for the lead convoy agent and independent agent. A proportional error controller is then used to actuate the acceleration of the aforementioned agents while adhering to a set of linear kinematic constraints. Meanwhile the tail convoy agent is actuated by a proportional error controller that aims to maintain a convoy distance over relative velocity of $2.24\ s$, with a $0.5\ s$ time lag to mimic a reasonable human reaction time. In both of these cases the proportional error controller in question calculates the vehicle acceleration by multiplying the velocity/distance error by a predefined gain parameter. For our experiments we found a proportional gain of $1.0$ sufficed to produce satisfactory causal scenes. The scenes were made to last for $50 - 70\ s$ and were generated at $10\ Hz$. Scenes were generated in such a way as to have $12$ causal interactions within the convoy and $12$ actions carried out by the independent agent. 


\subsection{Parameters}
The key parameters universal to all the methods were the significance alpha and maximum time lag, in order to avoid assessing the efficacy of the methods with a poor parameter selection these values were individually varied. The values used for the significance alpha were $0.001$, $0.005$, $0.01$, $0.03$, $0.05$ and $0.1$, while using a maximum time lag of $2.5\ s$ based upon a conservative estimate of reaction and actuation times of drivers. Meanwhile the values used for the maximum time lag were $2.5\ s$, $3.6\ s$ and $4.9\ s$, while using a significance alpha of $0.05$ based upon the previous work carried out by \citet{assaad2022survey}.

\subsection{Evaluation Metric}
To determine the efficacy of each evaluated method we calculate a True Positive (TP), False Positive (FP), and False Negative (FN) count for each causal scene. For a link within a discovered causal graph is a TP if present in the ground truth causal graph, and a false negative otherwise. Meanwhile, if a link present in the ground truth causal graph is not present in the discovered causal graph, this counts as a FN. The ground truth causal graph for all causal scenes we consider corresponds to the scenario described in Sec. \ref{sec:problem_definition} and illustrated in Fig. \ref{subfig:two_car_convoy_c}. From the previously calculated values we can derive the $\text{F}_1$ score for each causal scene and take the mean across scenes for a given method.
For additional details on the calculation of the TP, FP and FN counts please see the appendix.

\subsection{Results}

\begin{figure}[t]
    \centering
    \includegraphics[width=0.8\linewidth]{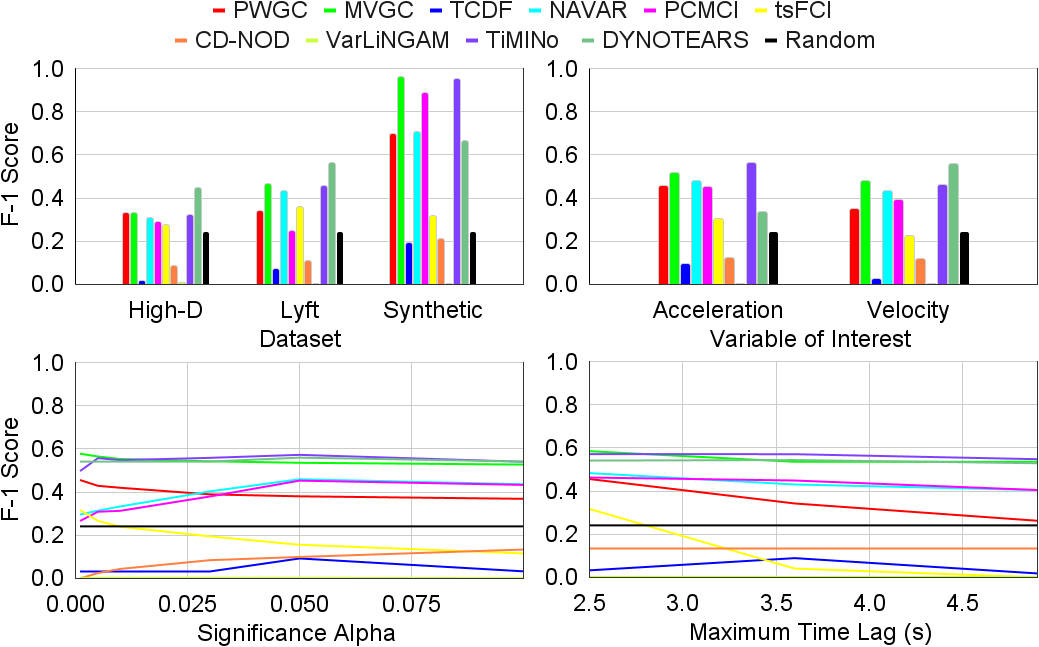}
    \caption{$\text{F}_1$ Score from applying each method to the Lyft, High-D and Synthetic datasets, for acceleration and velocity variables, while varying significance alpha and maximum time lag independently. For the dataset graph, the $\text{F}_1$ Score given for each method is based upon the optimal combination of parameters for the specific dataset-method combination. This process is similarly repeated in the other graphs, however the maximum $\text{F}_1$ Score is not taken across datasets in order to avoid the synthetic dataset dominating the figures. Instead the mean is taken across datasets, while still taking the optimal combination of parameters for parameters besides the parameter being varied.}
    \label{fig:f1_score}
\end{figure}


The results are displayed in Fig. \ref{fig:f1_score}. Overall the best mean $\text{F}_1$ scores are provided by DYNOTEARS, MVGC and TiMINo. PCMCI and NAVAR also provide competitive results, but are outperformed by at least one other method in every case. PWGC, tsFCI, and CD-NOD generally under-perform compared with other methods, while TCDF and VarLiNGAM completely fail in almost every case.

While it is possible that these failures are due to an issue of implementation, TCDF directly calls the same code utilised by the original paper \citep{nauta2019causal} and VarLiNGAM directly calls code from the Python \textit{lingam} package \citep{shimizu2011directlingam, hyvarinen2013pairwise}, making this unlikely. Theoretically speaking it is possible that TCDF is under-performing due to lack of training data, but this is made less likely by the fact NAVAR is competitive also training upon the same amount of data. Likewise with VarLiNGAM, it is possible that there is too weak of a direct coupling between variables for the noise of one variable to affect another, but considering TiMINo is competitive, this is again unlikely. Another possibility is that all variable noise is Gaussian, and TiMINo is still able to perform causal discovery despite this fact, while VarLiNGAM cannot.

In terms of parameter selection, significance alpha has the most influence although also the least consistent pattern of the two key parameters, making it an important parameter to tune for effective application of many methods. Interestingly DYNOTEARS appears to barely respond to changes in significance alpha, which could make it a better method to apply out of the box. Maximum time lag has a minimal impact across many methods, but in general over-estimating the maximum time lag leads to degradation in performance, thus a good approximation may suffice.

A clear issue illustrated by these results is the lack of readiness for these methods to be applied to real world data in these types of scenarios. While some methods such as MVGC and TiMINo are able to get close to a $\text{F}_1$ score of $1.0$ on synthetic data, the greatest performance we see on real world data is from DYNOTEARS at ${\sim}\,0.565$ $\text{F}_1$ score. This is clearly an unacceptable level of performance in a domain such as AD, demonstrating that either significant improvements need to be made to existing methodologies or alternate directions of research considered. This work does however highlight which methods might be candidates for further work, namely the aforementioned MVGC, TiMINo and DYNOTEARS methods.

Between real world data and synthetic data the main differences that could explain the degradation of performance are: increased levels of noise, causal non-stationarity and causal sparsity. Of these the latter two are of greater interest to us, as issues resulting from noise do not directly relate to causality research and can potentially be resolved at a lower level of abstraction (e.g. more accurate localisation). In terms of causal non-stationarity, the synthetic dataset effectively has stationarity as a result of the agent reaction time always being $0.5\ s$, meanwhile real world drivers feature inconsistent reaction times. As for causal sparsity, real world scenes we consider are typically $30\ s$ or under and contain a few causal interactions within the convoy at most. Meanwhile the synthetic scenes are $50-70\ s$ in length and contain $12$ causal interactions within the convoy. This highlights a key issue with attempting causal discovery in an online system where causal interactions may be infrequent and the observation window brief. The combination of these issues also likely explains the relatively poor performance of a method leveraging non-stationarity like CD-NOD. Here synthetic scenes provide ample yet mostly stationary data, while real scenes provide little in the way of data, that may or may not be non-stationary dependent upon whether more than a single causal interaction occurs.

\section{Discussion \& Conclusions}

The experiments establish that the evaluated approaches struggle with tackling even a simple causal scenario in the AD domain when working with real world data. However, these methods constitute the contemporary work of the temporal causal discovery field, indicating that further development of existing methodologies or new lines of thought entirely may be necessary to overcoming the new challenges presented within the AD domain.

The chief qualities of these new types of problems are causal non-stationarity and causal sparsity. The first of these has been identified in some recent literature as a matter of concern \citep{glymour2019review}. There has been some progress in this direction with one work applying a non-stationary causal discovery approach to medical data \citep{strobl2017causal} and another very recent work attempting causal discovery on conditionally stationary data \citep{rodas2021causal}. This latter example is interesting because it explores a physical causal relationship between particles through a spring, which more closely mirrors the types of relationships present in the AD domain. 

Causal sparsity as a quality is harder to overcome, as all observational approaches inherently rely upon evidence that is likely to be lacking when working in an online scenario-by-scenario fashion. It is here where a theoretical counterfactual approach to causal discovery might succeed by utilising approximations of individual agent behaviour to discover inter-agent causal relationships. While such an approach has been applied in the domain of climate science \citep{hannart2016causal}, to the author's knowledge no such research has been conducted in relation to autonomous agents.

Thus we suggest for future work that both of these avenues should undoubtedly be explored further, in doing so open up new avenues of causal reasoning between agents in autonomous robotics.


\acks{This work was supported by the EPSRC project RAILS (grant reference: EP/W011344/1) and the Oxford Robotics Institute research project RobotCycle.}


\bibliography{references.bib}

\appendix

\section{Data Preparation}

Because the real world datasets used in this paper are not open for public download we cannot share the data supplied as input for the experiments, only the data resulting from the experiments. However, in order to ensure the experiments are reproducible, we document the pre-processing steps taken on the datasets.

\subsection{Lyft Level 5 Prediction}

The first step in extracting scenes from the Lyft dataset is to extract the map information stored in the Protobuf format semantic map file using the ``extract\_map.py" script. Once this has been done, the agent data can be extracted via the ``extract\_agents.py" script.

At this point there are two options available, if the scenes can be visualised in some fashion\footnote{The tool utilised by the authors is part of another project and is not currently available to the public. However, the L5Kit provided by Lyft Level 5 on GitHub is quite useful for visualising scenes, and there is documentation of better visualisation methods available \citep{rojas2020autonomous}.} one can select scenes that encapsulate the two convoy scenario. From these scenes it is necessary to identify the ID number of the second convoy vehicle as well as choosing an independent vehicle. It is important to select an independent vehicle that is present in the scene for a majority of the time, as the scene will be trimmed to the time window where all agents are present. Furthermore it is also greatly preferable to select an independent agent that is not stationary for the entirety of the scene as to present a fair challenge to the methods under consideration. Once these details have been identified, ``convert\_to\_two\_agent\_followed\_scene.py" and ``convert\_to\_two\_agent\_follower\_scene.py" can be used to convert the extracted agent data to CSV time series files.

Alternatively, under the ``conversion" directory the ``agent\_json\_data\_to\_two\_agent\_convoy.sh" script will convert the selection of scenes and associated agent IDs used in the main paper to CSV time series files. Using this script requires setting the three environment variables, the input directory, output directory and the base Lyft dataset script directory (i.e. lyft\_prediction\_dataset\_tools).

\subsection{High-D}
With the additional information provided by the High-D dataset the process of scene extraction can be automated. Calling the ``extract\_two\_agent\_convoy\_scenes.py" script while passing the path to the High-D data directory will complete the whole extraction process. It is necessary to set a minimum scene length value and a minimum proportion velocity change for potential convoy agents, we selected $10\ s$ and $0.2$ respectively for these values.

\subsection{Synthetic}
For documentation of the steps to generate synthetic data, we refer the reader to the Git repository linked in Sec. \ref{subsec:datasets_and_code}. However, to summarise one can either call ``create\_two\_agent\_convoy\_scene.py" or ``create\_two\_agent\_convoy\_scenes.py" to generate a single or multiple causal scenes to be output to the specified file path or directory path respectively. Below the parameters used to generate the set of synthetic causal scenes used in this paper are documented.

\newpage
\begin{itemize}
    \item Variable: $\{ \text{``Acceleration"}, \text{``Velocity"} \}$
    \begin{itemize}
        \item Each of these parameter settings was used once independently to generate 100 acceleration-based causal scenes and 100 velocity-based causal scenes.
    \end{itemize}
    \item Frequency: $10.0\ Hz$
    \item Duration: $50.0 - 70.0\ s$
    \item Convoy Actions: $12$
    \item Independent Actions: $12$
    \item Minimum Convoy Distance: $10.0\ m$
    \item Maximum Convoy Distance: $100.0\ m$
    \item Proportional Coefficient: $1.0$
    \item Integral Coefficient: $0.0$
    \item Differential Coefficient: $0.0$
    \item Minimum Action Interval: $1.0\ s$
    \item Minimum Velocity: $0.0\ m/s$
    \item Maximum Velocity: $44.7\ m/s$
    \item Minimum Start Velocity: $4.47\ m/s$
    \item Maximum Start Velocity: $26.8\ m/s$
    \item Minimum Acceleration: $-6.56\ m/{s}^2$
    \item Maximum Acceleration: $3.5\ m/{s}^2$
    \item Safe Distance Over Velocity: $2.24\ s$
    \item Reaction Time $0.5\ s$
    \item Fixed Actuary Noise: $0.1 - 1.6\ m/{s}^2$
    \item Proportional Actuary Noise: $0.1 - 1.6$
    \item Fixed Sensory Noise: $0.01 - 0.16\ m$
    \item Proportional Sensory Noise: $0.005 - 0.08$
\end{itemize}

\section{Experiment Parameters} \label{sec:parameters}
Here we present the parameter values used by each method, and offer explanations for certain parameter choices. If an explanation is not offered it either is a default value based upon those utilised by previous works \citet{assaad2022survey,bussmann2021neural,zhang2017causal}, or one of the two parameters justified in the main paper (i.e. maximum time lag and significance alpha). We do not claim to assign optimal parameters to methods as optimising multiple parameters across 10 temporal causal discovery methods was deemed infeasible within the scope of the work. That being said, we would like to explore carrying out experiments for multiple significance alphas as part of future work.

\begin{itemize}
    \item Pairwise Granger
    \begin{itemize}
        \item Significance Alpha: $0.05$
        \item Statistical Test: Regression Sum of Squares (SSR) F-Test
        \item Maximum Time Lag: $25$ / $2.5\ s$
    \end{itemize}
    
    \item Multivariate Granger
    \begin{itemize}
        \item Significance Alpha: $0.05$
        \item Statistical Test: Chi-Squared
        \item Multiple Hypothesis Test Correction: Benjamini-Hochberg False Discovery Rate \citep{benjamini1995controlling}
        \item VAR Model Estimation Regression Mode: Ordinary Least Squares
        \item Information Criteria Regression Mode: Locally Weighted Regression \citep{cleveland1988locally}
        \item Model Order: Akaike Information Criterion \citep{akaike1973information}
        \item Maximum Time Lag: $25$ / $2.5\ s$
        \item Maximum Autocovariance Lags: $1000$
        \item Random Seed: Undefined
    \end{itemize}
    
    \item TCDF
    \begin{itemize}
        \item Significance Alpha: $0.05$
        \item Epochs: $1000$
        \item Learning Rate: $0.01$
        \item Hidden Layers: $1$
        \item Kernel Size: $5$
        \begin{itemize}
            \item While utilising a single hidden layer and identical values for kernel size and dilation coefficient, the maximum time lag consists of the kernel size/dilation coefficient value squared. Therefore in order to match the maximum time lag of 25 assigned to other methods we assign 5 to the kernel size/dilation coefficient here.
        \end{itemize}
        \item Dilation Coefficient: $5$
        \begin{itemize}
            \item See above.
        \end{itemize}
        \item Random Seed: $1111$
        \item Optimizer: Adam \citep{kingma2014adam}
        \item CUDA: False
        \begin{itemize}
            \item While CUDA usage might theoretically increase the speed of TCDF, we felt it best to avoid executing methods on the GPU where possible. This way the performance reflected in the main paper should act as a lower bound on the performance that can be improved with CUDA usage, rather than the reverse.
        \end{itemize}
    \end{itemize}

    \item NAVAR
    \begin{itemize}
        \item Significance Alpha: $0.05$
        \item Maximum Time Lag: $25$ / $2.5\ s$
        \item Hidden Nodes: $10$
        \item Hidden Layers: $1$
        \item Epochs: $2000$
        \item Batch Size: $32$
        \item Sparsity Penalty: $0.1$
        \item Weight Decay: $0.001$
        \item Dropout: $0.5$
        \item Learning Rate: $3.0 \times 10^{-4}$
        \item Validation Proportion: 0.0
        \item Network Type: Multi-Layer Perceptron
        \item Normalize: True
        \item Split Time Series: False
    \end{itemize}
    
    \item PCMCI
    \begin{itemize}
        \item Significance Alpha: $0.05$ (Used for both ``pc\_alpha" and ``alpha\_level")
        \item False Discovery Rate Method: Benjamini-Hochberg \citep{benjamini1995controlling}
        \item Minimum Time Lag: $0$ / $0.0\ s$
        \item Maximum Time Lag: $25$ / $2.5\ s$
        \item Maximum Number of Conditions to Test: Unrestricted
        \item Maximum Number of Conditions of Y to Use: Unrestricted
        \item Maximum Number of Conditions of X to Use: Unrestricted
        \item Maximum Number of Combinations of Conditions: $1$
        \item Conditional Independence Test: Partial Correlation \citep{baba2004partial}
        \begin{itemize}
            \item Partial correlation and conditional mutual information using K-nearest numbers were both used by \citet{assaad2022survey}. We tested both of these with the autonomous driving data, however the conditional mutual information approach took on average almost half an hour per run, making it intractable as an online causal discovery approach.
        \end{itemize}
    \end{itemize}
    
    \item tsFCI
    \begin{itemize}
        \item Significance Alpha: $0.05$
        \item Maximum Time Lag: $25$ / $2.5\ s$
        \item Include Instant Effects: False
        \item Data Type: Continuous
        \item Algorithm: tsCFCI (A variant of tsFCI with less stringent faithfulness requirements) \citep{ramsey2006adjacency}
    \end{itemize}

    \item CD-NOD
    \begin{itemize}
        \item Significance Alpha: $0.05$
        \item Maximum Time Lag: ?
        \begin{itemize}
            \item Although a maximum time lag parameter is described in the work discussing a time lagged relationship extension to CD-NOD \citep{huang2020causal} (i.e. $P$), there does not seem to be a way to specify this based upon the documentation available within the Git repository\footnote{https://github.com/Biwei-Huang/Causal-Discovery-from-Nonstationary-Heterogeneous-Data} nor from our attempts at interpreting the codebase.
        \end{itemize}
        \item Conditional Independence Test: Kernel-based Conditional Independence (KCI) \cite{zhang2011kernel}
        \item Surrogate Variable: Time Index
        \item Maximum Number of Conditioning Variables: 1
        \item Type: 0 (Run all phases)
        \item Pairwise: False
        \item Bonferroni Correction: False
        \item Conditional Independence Test GP Optimisation: True
        \item Direction Determination GP Optimisation: True
        \item Observational Variable Kernel Width: 0 (Automatically calculated via GP optimisation)
        \item Time Index Kernel Width: 0.1
    \end{itemize}
    
    \item VarLiNGAM
    \begin{itemize}
        \item VAR Model Estimation Regression Mode: Ordinary Least Squares
        \item Trend Assumption: Co-Constant, No Trend
        \item Pruning: True
        \item Significance Alpha: $0.05$
        \item Maximum Time Lag: $25$ / $2.5\ s$
        \item Regularisation Criterion: Bayesian Information Criterion \citep{schwarz1978estimating}
        \item Random Seed: Undefined
        \item Algorithm: DirectLiNGAM (A variant of LiNGAM which typically executes faster) \citep{shimizu2011directlingam}
    \end{itemize}
    
    \item TiMINo
    \begin{itemize}
        \item Significance Alpha: $0.05$
        \item Maximum Time Lag: $25$ / $2.5\ s$
        \item Assumed Time Series Model: Linear
        \item Independence Test: Cross Covariance
        \item Include Instant Effects: False
        \item Check for Confounders: False
    \end{itemize}
    
    \item DYNOTEARS
    \begin{itemize}
        \item Threshold for W: $0.01$
        \item Threshold for A: $0.01$
        \item Regularisation Constant for W: $0.05$
        \item Regularisation Constant for A: $0.05$
        \item Maximum Time Lag: $25$ / $2.5\ s$
        \item Maximum Number of Iterations: $100$
        \item Acyclicity Tolerance: $1.0 \times 10^{-8}$
    \end{itemize}
    
    \item Random Causal Discovery
    \begin{itemize}
        \item Edge Likelihood: $0.5$
        \begin{itemize}
            \item This was selected purely based upon it being the most naive approach to randomly constructing a graph (i.e. Flipping a coin for each potential edge effectively) as the intended use for this method was as a baseline.
        \end{itemize}
    \end{itemize}
\end{itemize}


\newpage
\section{Experiment Setup} \label{sec:experiment_setup}

\subsection{Hardware}

\begin{itemize}
    \item CPU: AMD Ryzen 9 3950X
    \item GPUs: Nvidia GeForce RTX 2070 SUPER, Nvidia GeForce GTX 750 Ti
    \item Storage: Samsung Electronics 970 EVO Plus NVMe M.2 Internal SSD
\end{itemize}

\subsection{Software}

\begin{itemize}
    \item Kernel: Linux version 5.4.0-137-generic
    \item OS: Ubuntu 20.04.5 LTS ``Focal"
    \item GCC/G++: 10.3.0
    \item Python: 3.8.10
    \begin{itemize}
        \item numpy: 1.23.3
        \item pandas: 1.4.3
        \item scikit-learn: 0.23.1
        \item scipy: 1.4.1
        \item statsmodels: 0.11.1
        \item joblib: 0.15.1
        \item graphviz: 0.8.4
        \item networkx: 2.8.5
        \item matplotlib: 3.1.2
        \item torch: 1.12.1+cu113
        \item l5kit: 1.1.0
    \end{itemize}
    \item R: 4.2.2 ``Innocent and Trusting"
    \item Matlab: R2022a Update 4 (9.12.0.2009381)
\end{itemize}


\newpage
\section{Experiment Process}

Carrying out a set of experiments corresponds to setting the significance alpha and maximum time lag and carrying out a number of runs of ``test\_ad.py" while varying three parameters:
\begin{itemize}
    \item Method: GrangerPW, GrangerMV, TCDF, NAVARMLP, PCMCIParCorr, tsFCI, CDNOD, VarLiNGAM, TiMINo, Dynotears
    \item Dataset: lyft, highd, synthetic
    \item Variable: acceleration, velocity
\end{itemize}
The first two of these correspond to the method to use and the dataset to apply it to. The third parameter specifies a sub-dataset in the sense that it determines whether or not to use acceleration or velocity as the variable of interest for agents.

Each run of ``test\_ad.py" applies the selected method to every scene in the sub-dataset, before checking the discovered causal graphs against the predetermined ground truth using the true positive, false positive, and false negative metrics defined in the main paper. The final output from the run is all of the graphs discovered for each scene, the individual $\text{F}_1$ Score, Precision, Recall and Runtime for each scene, and the mean and standard deviation of these metrics across all scenes.

In order to calculate $\text{F}_1$ Score, Precision and Recall, we calculated the True Positive (TP), False Positive (FP) and False Negative (FN) count for the $i$-th scenario as follows:
\begin{equation}
\small
    |{TP}_i| = \sum_{v^j \in V} | \hat{pa}_i (v^j) \cap pa(v^j) |
\end{equation}
\begin{equation}
\small
	|{FP}_i| = \sum_{v^j \in V} | \hat{pa}_i (v^j) \setminus pa(v^j) |
\end{equation}
\begin{equation}
\small
	|{FN}_i| = \sum_{v^j \in V}  | pa(v^j) \setminus \hat{pa}_i (v^j) |
\end{equation}
where $pa(v^j)$ represents the ground truth scenario independent parents for the time series $v^j$ and $\hat{pa}_i (v^j)$ represents the parents discovered for the $i$-th scenario by applying a given method. Although some methods can detect self-causal relationships, these have no clear meaning in the context of agent behavioural interaction, and therefore these relationships are excluded from the evaluation.

\end{document}